\PassOptionsToPackage{numbers,sort&compress,square}{natbib}
\documentclass{article}
\usepackage{amssymb}
\usepackage{float}
\usepackage{booktabs}
\usepackage{amssymb}
\usepackage{graphicx}
\usepackage{amsmath}
\usepackage{wrapfig}
\usepackage{caption}
\usepackage{color}
\usepackage{makecell}
\usepackage{pifont} 
\usepackage{tikz}   
\usepackage{bbm}

\usepackage{graphicx}
\usepackage{mwe} 
\usepackage{multirow}
\usepackage{makecell}  
\usepackage{xcolor}
\usepackage{pifont}
\definecolor{turquoise}{cmyk}{0.65,0,0.1,0.3}
\definecolor{purple}{rgb}{0.65,0,0.65}
\definecolor{dark_green}{rgb}{0, 0.5, 0}
\definecolor{orange}{rgb}{0.8, 0.6, 0.2}
\definecolor{red}{rgb}{0.8, 0.2, 0.2}
\definecolor{darkred}{rgb}{0.6, 0.1, 0.05}
\definecolor{blueish}{rgb}{0.0, 0.3, .6}
\definecolor{light_gray}{rgb}{0.7, 0.7, .7}
\definecolor{pink}{rgb}{1, 0, 1}
\definecolor{greyblue}{rgb}{0.25, 0.25, 1}
\definecolor{brinkpink}{rgb}{0.98, 0.38, 0.5}
\definecolor{codegreen}{rgb}{0,0.6,0}
\definecolor{codegray}{rgb}{0.5,0.5,0.5}
\definecolor{codepurple}{rgb}{0.58,0,0.82}
\definecolor{backcolour}{rgb}{0.95,0.95,0.92}

\usepackage[utf8]{inputenc}
\usepackage{tcolorbox}
\tcbuselibrary{breakable}
\tcbuselibrary{listings,skins}
\usepackage{listings}
\newcommand{\ourmethod}{\textsc{DexFLEX}}
\newcommand{\equaladvising}{%
  \textsuperscript{\textnormal{\textdagger}}%
}

\usepackage[final]{corl_2026} 

\title{DexFLEX: Contact-Aware Foundation Controller for Command-Guided Dexterity}

\author{
\textbf{
Weikang Wan\textsuperscript{1},
Zeyu Lai\textsuperscript{1},
Jianyu Hou\textsuperscript{1},
Yuchen Zhou\textsuperscript{2},
}\\
\textbf{
Jiawei Fu\textsuperscript{1},
Henrik I.\ Christensen\textsuperscript{1}\equaladvising,
Hao Su\textsuperscript{1,3}\equaladvising
}\\[0.5em]
{\textsuperscript{1}University of California San Diego
\quad
\textsuperscript{2}University of Pennsylvania
\quad
\textsuperscript{3}Sudo AI GmbH
}
}

\begin{document}
\maketitle

\begingroup
\renewcommand{\thefootnote}{\fnsymbol{footnote}}
\footnotetext[2]{Equal advising.}
\endgroup

\vspace{-5mm}
\begin{abstract}
Dexterous robot hands can receive useful motion intent from teleoperation or learned policies, but successful execution also depends on contact decisions that these commands rarely specify.
We introduce \ourmethod{}, a contact-aware foundation controller that turns upstream fingertip-motion drafts into contact-consistent joint commands.
Instead of treating a draft as a trajectory to copy, \ourmethod{} treats it as evidence about intent: from the current tactile--proprioceptive state, it proposes feasible short-horizon motion chunks, predicts their contact consequences, and selects the candidate that best follows the command while preserving future contact stability.
At inference time, a simple \emph{Draft--Dream--Select} procedure combines pure-prior proposals for recovery with draft-seeded proposals for responsiveness, then decodes the selected motion into executable joint-space targets.
Across degraded-command simulation, real-world shared control, and visuomotor policy learning, the same trained controller improves robustness without retraining, increasing real-world teleoperation success from 29.2\% to 78.3\% and reaching 46.7\% policy-learning success, 3.5$\times$ direct joint-action prediction, and 1.56$\times$ prior-only correction.
Videos are available
at \href{https://dex-flex.github.io/}{dex-flex.github.io}.
\end{abstract}

\vspace{-3mm}
\begin{figure}[!h]
    \centering
    \includegraphics[width=\textwidth]{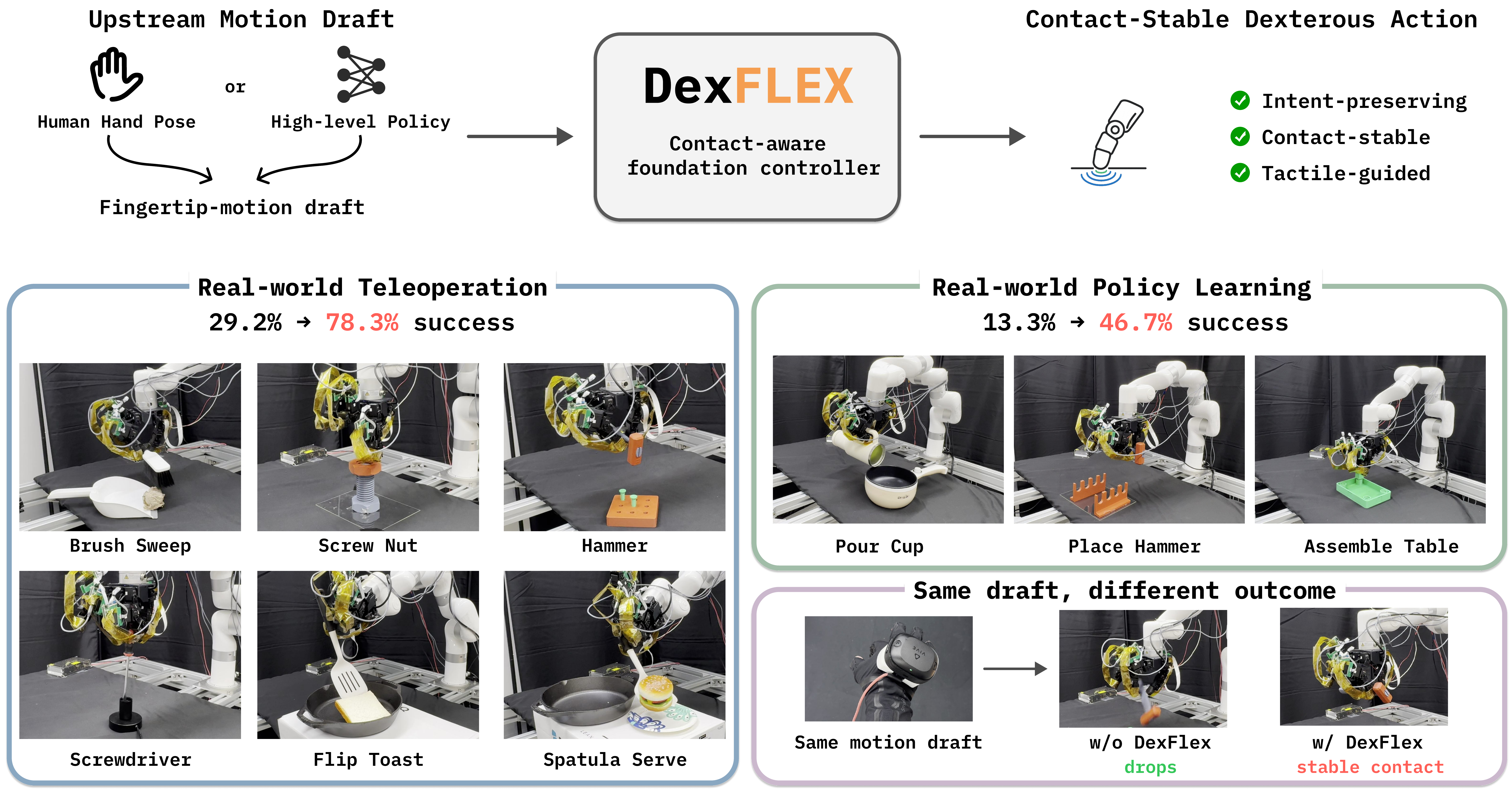}
    \caption{\textbf{\ourmethod{} converts upstream motion drafts into contact-stable dexterous actions.}
    Human operators or learned policies provide fingertip-motion drafts, and \ourmethod{} supplies tactile-guided low-level correction for contact-consistent execution.
    With the same draft, \ourmethod{} preserves contact where direct execution can fail.}
    \label{fig:pull-figure}
\end{figure}
\vspace{-2mm}

\section{Introduction}
\label{sec:intro}
\vspace{-2mm}
Dexterous robot hands can receive useful motion intent from teleoperation, retargeting, or learned policies, but successful execution also depends on contacts that are rarely specified by those commands.
A short fingertip-motion draft may indicate where the fingertips should move next, yet omit which contacts should remain load-bearing, which fingers should unload, or when support should transfer during regrasping and tool use.
As a result, a kinematically reasonable motion can still fail by releasing a support finger, breaking contact too early, or disturbing a stable grasp.
We study this problem as \emph{command-guided dexterity}: a low-level controller must preserve useful upstream intent while making execution contact-consistent.

Existing methods address only parts of this problem.
Shared-autonomy and generative dexterous-control methods can refine user or policy commands with learned motion priors~\cite{reddy2018shared,schaff2020residual,yoneda2023noise,yin2025dexteritygen}, but prior-only correction mainly evaluates motion plausibility or draft agreement, not the contact outcome a motion will produce.
Two fingertip motions can be similarly close to the draft, while only one preserves a supporting contact or stable tool grasp.
Tactile and visuotactile policies show that touch is crucial for contact-rich dexterity~\cite{yin2023rotating,qi2023general,guzey2023dexterity}, yet they are typically trained end-to-end for specific tasks rather than as reusable low-level controllers beneath different command sources.
This motivates a contact-aware foundation controller: a reusable dexterous executor guided by motion commands, grounded in tactile state, and compatible with multiple upstream interfaces.

We introduce \ourmethod{}, a \emph{Contact-Aware Foundation Controller for Command-Guided Dexterity}.
The key idea is to treat the upstream draft not as a trajectory to copy, but as online evidence about intent.
\ourmethod{} follows a propose--predict--select principle: a state-conditioned consistency-flow prior proposes short-horizon fingertip-motion chunks from tactile--proprioceptive state, a latent world model predicts each candidate's contact consequence, and a learned score selects the candidate that best preserves the draft while avoiding contact failures.
At inference, \ourmethod{} implements this through \emph{Draft--Dream--Select}, mixing pure-prior proposals for recovery with draft-seeded proposals for responsiveness before decoding the selected chunk into joint-space targets.
This keeps the learned motion prior reusable while allowing the controller to repair contact-critical details without simply damping the command or overriding upstream intent.

We evaluate \ourmethod{} in degraded-command simulation, real-world shared control, and visuomotor policy learning.
In simulation, we corrupt expert commands with noise, scaling, latency, and support-finger removal to test intent recovery under contact constraints.
In the real world, operators complete six long-horizon tool-use tasks, and an image-based flow matching policy uses \ourmethod{} as a frozen low-level action interface.
\ourmethod{} improves average real-world success from 29.2\% with raw teleoperation to 78.3\%, outperforming DexGen's 52.5\% by 25.8 percentage points.
As a low-level policy interface, it reaches 46.7\% average success, \textbf{3.5}$\times$ the success rate of direct joint-action prediction and \textbf{1.56}$\times$ that of prior-only correction.

\noindent\textbf{Contributions.}
Our contributions are: (1) we formulate command-guided dexterity as contact-aware low-level refinement of fingertip-motion drafts; (2) we introduce \ourmethod{}, a reusable controller that combines state-conditioned motion proposals, latent consequence prediction, and Draft--Dream--Select execution; and (3) we show that the same trained module improves degraded-command recovery, real-world shared control, and visuomotor policy learning without retraining.

\section{Related Work}
\label{sec:related}
\vspace{-2mm}
\textbf{Foundation controllers and motion priors for dexterous manipulation.}
Recent progress in robot foundation models has explored large-scale pretraining as a way to obtain generalist policies across tasks and environments~\citep{brohan2022rt,zitkovich2023rt,pmlr-v305-black25a}.
Most target end-to-end visuomotor control for broad manipulation, whereas contact-rich dexterous hands require low-level execution that responds to fine-grained motion and contact changes~\citep{lee2020making,xu2023unidexgrasp,ye2025learning,wan2025lodestar,zheng2026omnivta, higuera2026visuo, yuan2026vtam,niu2026learning}.
Teleoperation and retargeting provide natural upstream interfaces for such fine-grained dexterous guidance~\citep{handa2020dexpilot,qin2023anyteleop,wang2024dexcap,jiang2025dexmimicgen,pan2025spider}.
Most closely related, DexGen learns a teleoperation-prompted dexterous foundation controller for generating robot-hand actions~\citep{yin2025dexteritygen}.
\ourmethod{} shares the goal of reusable dexterous control, but separates the state-conditioned motion prior from command-dependent correction: the prior proposes feasible fingertip-motion chunks independent of the command source, while the draft is used to select among candidates.
By selecting motions using predicted contact consequences rather than draft agreement alone, \ourmethod{} serves as a contact-aware low-level executor beneath teleoperation, retargeting, and learned policies.

\paragraph{Shared autonomy and tactile contact correction.}
Shared autonomy improves imperfect user commands through intent inference, policy blending, residual correction, safety filtering, or learned command refinement~\cite{aarno2005adaptive, kim2006continuous, schroer2015autonomous, schwarting2017parallel,dragan2013policy,javdani2018shared,reddy2018shared,schaff2020residual,yoneda2023noise}.
Dexterous manipulation adds a contact-level ambiguity: two motions can match the user's intent, while only one preserves supporting contacts or avoids premature release.
Tactile and visuotactile methods show that touch is critical for resolving such ambiguities in in-hand rotation, contact-rich control, and dexterous policy learning~\cite{yin2023rotating,qi2023general,guzey2023dexterity,lin2025learning,suresh2024neuralfeels,xu2026fingereye}, but are usually trained as task-specific policies or representation learners.
\ourmethod{} combines shared-autonomy-style intent preservation with tactile contact reasoning by treating the upstream draft as intent evidence and selecting candidate fingertip motions using predicted contact consequences.
\begin{figure}[t]
    \centering
    \includegraphics[width=1.0\textwidth]{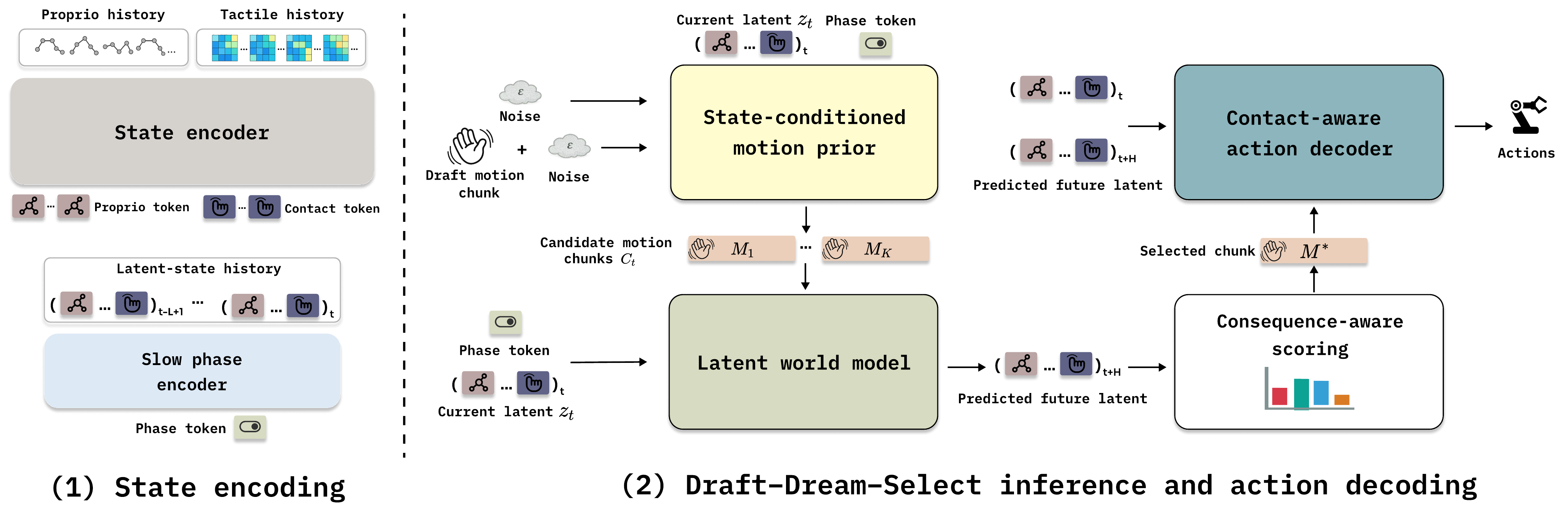}
    \caption{\textbf{\ourmethod{} pipeline.}
    \textbf{(1)} A state encoder converts proprioceptive and tactile histories into a contact--proprioceptive latent $z_t$, whose temporal history is summarized as a slow phase token $g_t$.
    \textbf{(2)} At inference, Draft--Dream--Select proposes pure-prior and draft-seeded motion chunks, predicts their contact consequences, scores them against the upstream draft, and decodes the selected chunk into joint-space actions.}
    \label{fig:pipeline}
    \vspace{-5mm}
\end{figure}

\section{Method}
\label{sec:method}
\vspace{-3mm}
\textbf{Problem setup.}
At each control step $t$, the robot receives recent proprioception $o_t$ and fingertip tactile observations $\mathcal{T}_t$.
An upstream interface---such as glove teleoperation, retargeting, or a learned high-level policy---provides a short fingertip-space motion draft
$\tilde M_t=\{\Delta \tilde x_{t+i}\}_{i=1}^{H}$, where $x_t$ denotes the current fingertip keypoints and $H$ is a short horizon.
The role of the low-level controller is to convert this draft into a contact-consistent motion chunk
$M_t=\{\Delta x_{t+i}\}_{i=1}^{H}$ that preserves the intended fingertip motion while avoiding contact failures such as premature release or loss of support.
The selected chunk is then decoded into a joint-space target $a_t$ for execution.
We use fingertip-space motion as the shared upstream interface because it avoids exposing embodiment-specific joint coordinates to teleoperation or policy modules, while still providing enough geometric structure for dexterous manipulation.

\textbf{Algorithm overview.}
\ourmethod{} is a reusable low-level controller for this refinement problem (Fig.~\ref{fig:pipeline}).
Its design follows a simple propose--predict--select principle.
First, a state-conditioned motion prior proposes feasible short-horizon fingertip motions from the current tactile--proprioceptive state, independent of where the draft comes from.
Second, a latent world model predicts the contact--proprioceptive consequence of each candidate.
Third, a learned score selects the candidate that best preserves the upstream draft while avoiding contact failures, and an action decoder converts the selected chunk into a joint-space target.
We refer to this procedure as command-conditioned posterior correction: the prior defines the feasible motion support, while command dependence enters through consequence-aware scoring over a finite candidate set.
We describe \ourmethod{} in three parts: state-conditioned motion proposals (Sec.~\ref{sec:prior}), consequence-aware scoring (Sec.~\ref{sec:score}), and Draft--Dream--Select execution with action decoding (Sec.~\ref{sec:inference}).

\subsection{State-Conditioned Motion Proposals}
\label{sec:prior}

\textbf{Contact--proprioceptive state.}
\ourmethod{} encodes recent proprioception and fingertip tactile observations into a controller state $s_t=(z_t,g_t)$.
The continuous latent $z_t=[p_t,c_t]$ combines a proprioceptive token $p_t$ with a structured contact token $c_t$.
To preserve finger-level interaction structure, each fingertip taxel field is encoded separately and fused with finger-local proprioceptive features; the resulting per-finger tokens are aggregated with a lightweight attention module and a global contact token.
This structure preserves information about which fingers are carrying load, being unloaded, or likely to transfer support, instead of collapsing all tactile readings into a single flattened vector.
We also infer a low-rate discrete latent $g_t$ from a longer state window, intended to capture slowly varying manipulation regimes such as holding, pivoting, regrasp preparation, and release.
The encoder is trained jointly with the motion prior and then kept fixed for consequence prediction and scoring.

\textbf{Dexterous motion prior.}
Given $s_t$, the motion prior proposes short-horizon fingertip-motion chunks that are consistent with the current contact state, independent of the upstream draft.
We instantiate the prior as a state-conditioned consistency-flow model~\cite{yan2025maniflow}, which provides expressive proposals with few function evaluations for online control.
The flow network uses a lightweight transformer conditioned on the structured state tokens $\{p_t,c_t,g_t\}$ through cross-attention, allowing proposal generation to depend on both hand configuration and per-finger contact context.
The prior is trained on expert motion chunks from the demonstration dataset $\mathcal{D}$ using flow matching with a shortcut-consistency loss; details are given in Appendix.
At inference, it is used only as a proposal generator: it defines the feasible motion support from which \ourmethod{} later selects using the upstream draft and predicted contact consequence.
We apply tactile conditioning dropout during training to improve robustness to missing or noisy tactile readings.

\subsection{Consequence-Aware Scoring}
\label{sec:score}
The state-conditioned prior can produce many feasible motion chunks, but feasibility alone does not determine which chunk should be executed under the current draft.
Two candidates may be similarly plausible and similarly close to $\tilde M_t$, while only one preserves a supporting contact or avoids premature release.
\ourmethod{} resolves this ambiguity by predicting the short-horizon consequence of each candidate and scoring candidates using both draft alignment and predicted contact outcome.

\textbf{Latent world model.}
Given the current controller state $s_t$ and a candidate chunk $M$, a latent world model $F_\omega$ predicts the terminal contact--proprioceptive latent $\hat z_{t+H}=F_\omega(s_t,M)$ after executing the chunk.
Because the horizon is short, we keep the slow discrete latent $g_t$ fixed during this dreamed rollout and predict only the continuous latent.
The model is trained by latent regression in simulation using expert chunks, prior samples, and locally perturbed expert chunks.
For each candidate, the target latent is obtained by replaying the chunk from the corresponding simulator state and encoding the resulting tactile--proprioceptive observation.
Training on these candidate sources keeps the consequence model calibrated for the motions it will evaluate during selection, rather than only for expert demonstrations.

\textbf{Command- and consequence-aware score.}
The score $S_\eta(s_t,\tilde M_t,M,F_\omega(s_t,M))$ assigns a scalar value to each candidate.
It uses $s_t$ to interpret the current contact context, $\tilde M_t$ as evidence of upstream intent, $M$ as the candidate motion, and $F_\omega(s_t,M)$ as the candidate's predicted consequence.
This allows the controller to prefer a candidate that slightly modifies the draft to preserve support over one that matches the draft more closely but is predicted to break contact.

Since the dataset contains expert state--motion--action tuples but no paired examples of imperfect drafts and desired corrections, we train the score with synthetic proxy drafts.
For an expert chunk $M_t$, we construct $\tilde M_t=\mathcal{A}_{\mathrm{cmd}}(M_t)$ using mild masking, small perturbations, and coarse temporal corruption.
The score is trained as a contrastive ranker: the expert chunk is the positive candidate, while negatives include prior samples, hard negatives from other states with similar proxy drafts, and proposal-matched negatives from the draft-seeded branch.
This training uses proxy drafts only as noisy intent cues; at deployment, the candidate set is still generated from the state-conditioned prior, and the real upstream draft enters only through scoring.
Implementation details and negative-mix ablations are provided in Appendix.

\subsection{Draft--Dream--Select Execution}
\label{sec:inference}
At deployment, \ourmethod{} executes a simple Draft--Dream--Select loop.
Given the current state $s_t$ and upstream draft $\tilde M_t$, it forms a finite candidate set from two proposal branches, predicts each candidate's consequence with $F_\omega$, and selects the highest-scoring chunk:
\begin{equation}
\label{eq:dds}
\mathcal{C}_t=\mathcal{C}^{\mathrm{pure}}_t\cup\mathcal{C}^{\mathrm{draft}}_t,
\qquad
M_t^\star=\arg\max_{M\in\mathcal{C}_t}
S_\eta\big(s_t,\tilde M_t,M,F_\omega(s_t,M)\big).
\end{equation}
The pure-prior branch samples from the state-conditioned prior and provides recovery options when the draft is unsafe or incomplete.
The draft-seeded branch initializes the same flow from a noised version of $\tilde M_t$ at an intermediate noise level, providing local refinements that remain responsive to the upstream command.
Importantly, the flow model itself is still conditioned only on $s_t$; the draft changes the proposal initialization used for finite-candidate search, not the learned motion prior.

\textbf{Action decoding.}
The selected fingertip-motion chunk is converted into an executable joint-space target by an inverse-dynamics decoder,
$a_t=I_\phi(s_t,M_t^\star,F_\omega(s_t,M_t^\star))$.
Conditioning the decoder on the dreamed consequence helps disambiguate cases where similar fingertip displacements require different joint realizations, such as consolidating a grasp, unloading a support finger, or entering a release transition.
At runtime, selection is performed in a receding-horizon manner: \ourmethod{} executes a short prefix of the selected chunk, updates the tactile--proprioceptive state, and then repeats Draft--Dream--Select.
Timing, proposal counts, and control-frequency details are provided in Appendix.

\textbf{Training and reuse.}
We train the components sequentially to keep each interface stationary: the encoder and motion prior are trained first, followed by the consequence model, the score, and finally the action decoder.
After training, the same low-level module can be reused with any upstream source that provides fingertip-motion drafts in the same command space.
In shared control, the draft comes from glove teleoperation or retargeting; in policy learning, a high-level policy predicts the draft and \ourmethod{} executes it as a frozen contact-aware low-level controller.

\section{Experiments}
\label{sec:exp}
\vspace{-2mm}
We evaluate \ourmethod{} as a reusable contact-aware low-level controller through three questions.
\textbf{Q1:} Can it correct degraded upstream commands while preserving intended motion?
\textbf{Q2:} Can it assist human users in real-world long-horizon tool use?
\textbf{Q3:} Can it serve as a frozen low-level interface for visuomotor policy learning?
Together, these settings test whether the same learned controller can preserve upstream intent while repairing contact-critical execution details.
Unless otherwise noted, \ourmethod{} is trained once and reused across settings without retraining.

\subsection{Experimental Setup}
\label{sec:exp_setup}
\noindent\textbf{Training data.}
We train \ourmethod{} from simulated tactile--proprioceptive rollouts in Isaac Gym.
The data is collected with a 16-DoF LEAP four-fingered hand or a 22-DoF SHARPA five-fingered hand, each mounted on a 7-DoF xArm, over large-scale procedurally generated objects.
RL experts trained on an Any-Goal-Pose Reaching task~\citep{kedia2026simtoolreal} provide expert fingertip-motion chunks and joint targets.
This single dataset trains all learned components of \ourmethod{}; rollout scale, filtering, and training details are provided in Appendix.

\noindent\textbf{Real-world platform.}
For shared-control experiments, we use a 16-DoF LEAP hand mounted on a 7-DoF xArm7.
Human finger motion is captured by a Rokoko Smart Glove, and wrist pose is tracked by a Vive Ultimate tracker.
The teleoperation stack retargets human motion to fingertip and wrist targets and solves inverse kinematics with Mink~\citep{mink} at 50\,Hz.
Each fingertip is equipped with a $10\times6$ FlexiTac tactile sensor~\citep{huang20243d,huang2026flexitac}, whose spatial taxel layout is used by the structured contact encoder in Sec.~\ref{sec:prior}.
Sensor calibration, synchronization, and inference-time proposal hyperparameters are provided in Appendix.

\noindent\textbf{Baselines and metrics.}
For command-correction experiments, we compare against \textsc{Raw}, which directly executes the upstream draft; \textsc{Filter}, which applies low-pass filtering and command clipping; and \textsc{DexGen}, a diffusion-based prior-only correction baseline~\cite{yin2025dexteritygen} without latent consequence prediction or consequence-aware scoring.
We also include \ourmethod{} \textsc{w/o Tactile}, which removes tactile/contact tokens while keeping the rest of the pipeline unchanged.
In simulation, we report normalized rollout duration, task progress, and intent recovery error.
In real-world teleoperation, we report success rate (SR), completion time over successful trials (CT), and normalized holding time (NHT), the fraction of the task horizon with a stable functional grasp.
Each real-world teleoperation task is evaluated over 20 trials per method across two operators after a 30-minute per-interface familiarization period.
For policy learning, all methods use 50 demonstrations per task and the same image-based flow matching policy backbone~\cite{lipman2022flow}, varying only the low-level action interface.

\subsection{Recovering Intent from Degraded Commands}
\label{sec:sim_robustness}

\begin{figure}[t]
\centering
\includegraphics[width=\textwidth]{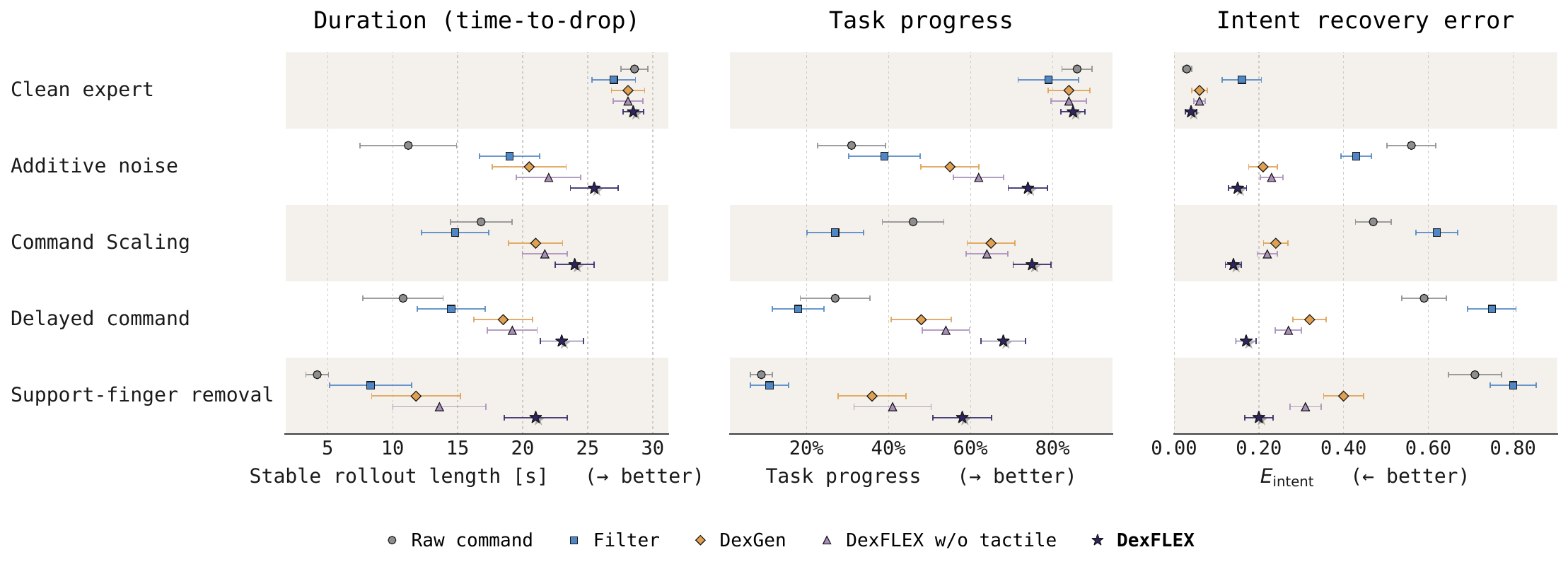}
\vspace{-3mm}
\caption{\textbf{Robustness under command degradation.}
Expert commands are corrupted with additive noise, command scaling, delay, and support-finger removal. We evaluate whether each
controller preserves contact stability while recovering the expert intent rather than merely damping motion.}
\label{fig:robustness_sim}
\vspace{-5mm}
\end{figure}

\begin{wraptable}[14]{r}{0.46\textwidth}
\vspace{-1\baselineskip}
\centering
\footnotesize
\setlength{\tabcolsep}{3.5pt}
\begin{tabular}{@{}lccc@{}}
    \toprule
    \textbf{Variant}
    & \textbf{Dur.} $\uparrow$
    & \textbf{Prog.} $\uparrow$
    & \textbf{Intent} $\downarrow$ \\
    \midrule
    \ourmethod{}
    & \textbf{0.86} & \textbf{0.78} & \textbf{0.18} \\
    \quad w/o predictive score
    & 0.52 & 0.46 & 0.45 \\
    \quad w/o tactile/contact
    & 0.62 & 0.55 & 0.36 \\
    \quad w/o consequence pred.
    & 0.64 & 0.57 & 0.34 \\
    \quad w/o pure-prior prop.
    & 0.66 & 0.61 & 0.25 \\
    \quad w/o draft-seeded prop.
    & 0.74 & 0.66 & 0.38 \\
    \quad w/o slow phase token
    & 0.81 & 0.74 & 0.22 \\
    \bottomrule
\end{tabular}
\caption{\textbf{Component ablations.}
Metrics are averaged over the four command-degradation settings. \emph{w/o predictive score} replaces the learned score with nearest-to-draft selection over the same proposal set.}
\label{tab:component_ablation}
\vspace{-1mm}
\end{wraptable}

A useful low-level controller should not simply suppress unsafe commands; it should recover the intended motion when the draft is noisy, delayed, or incomplete, while intervening when the draft would destabilize contact.
Simulation provides clean expert commands $u_t^\star$, allowing us to measure recovery against an external reference.
Starting from clean expert commands $u_t^\star$, we construct four degraded command streams: 
\textsc{Additive Noise}, $\tilde u_t=u_t^\star+\xi_t$ with $\xi_t\sim\mathcal U(-\alpha,\alpha)$; 
\textsc{Command Scaling}, $\tilde u_t=\beta_t u_t^\star$; 
\textsc{Delayed Command}, $\tilde u_t=u_{t-\Delta}^\star$; 
and \textsc{Support-Finger Removal}, which replaces the commands of load-bearing fingertips with release-like motions.
We report normalized rollout duration, task progress, and intent recovery error
$E_{\mathrm{intent}}=\frac{1}{T}\sum_t\frac{\|\hat u_t-u_t^\star\|_2}{\|u_t^\star\|_2+\epsilon}$,
where $\hat u_t$ is the executed command after low-level correction.
Lower intent error means the controller recovers the clean expert command rather than merely staying close to the corrupted draft.
Fig.~\ref{fig:robustness_sim} shows that \ourmethod{} improves all three metrics across command degradations.
The largest gains occur under \textsc{Support-Finger Removal}, where \textsc{Raw} often drops the object by removing stabilizing contacts, while \ourmethod{} selects motions that preserve support and still follow the command direction.
The gains are not explained by conservative damping: \textsc{Filter} can improve duration by suppressing motion, but increases intent error and reduces task progress.
\textsc{DexGen} improves over simple baselines through prior-based correction, and \ourmethod{} \textsc{w/o Tactile} improves further through consequence-aware selection, but the full model performs best under contact-critical corruptions by using tactile state to identify load-bearing fingertips.

\begin{table*}[t]
\centering
\resizebox{\textwidth}{!}{
\begin{tabular}{lccccccccc}
\toprule
\textbf{Method}
& \textbf{Brush} & \textbf{Nut} & \textbf{Hammer} & \textbf{Screwdrv.} & \textbf{Spatula} & \textbf{Toast}
& \textbf{Avg SR} $\uparrow$ & \textbf{Avg CT} $\downarrow$ & \textbf{Avg NHT} $\uparrow$ \\
\midrule
Raw Teleop
& 50 & 30 & 40 & 30 & 15 & 10
& 29.2 & 27.5 & 0.42 \\
Filter
& 60 & 40 & 50 & 40 & 25 & 20
& 39.2 & 33.8 & 0.55 \\
DexGen
& 70 & 55 & 60 & 55 & 40 & 35
& 52.5 & 31.2 & 0.65 \\
\ourmethod{} w/o Tactile
& 75 & 60 & 65 & 60 & 50 & 45
& 59.2 & 30.8 & 0.71 \\
\ourmethod{}
& \textbf{90} & \textbf{80} & \textbf{85} & \textbf{80} & \textbf{70} & \textbf{65}
& \textbf{78.3} & 32.0 & \textbf{0.85} \\
\bottomrule
\end{tabular}
}
\caption{\textbf{Real-world teleoperation results.}
We report per-task success rate (SR, \%) and average SR / CT / NHT across tasks. CT is completion time in seconds over successful trials. NHT is normalized holding time. Each method is evaluated for 20 trials per task across two operators. Averages are computed over task-level means.}
\label{tab:real_teleop_results}
\vspace{-5mm}
\end{table*}

\noindent\textbf{Component ablations.}
Table~\ref{tab:component_ablation} isolates the main design choices.
Removing consequence-aware scoring causes the largest drop, showing that selection over prior samples is not reducible to nearest-to-draft matching.
Removing tactile input or consequence prediction also degrades performance, especially under contact-critical corruptions, confirming that both current contact state and predicted contact outcome are needed for robust correction.
The proposal branches are complementary: pure-prior proposals provide recovery options when the draft is unsafe, while draft-seeded proposals preserve responsiveness to upstream intent.

\subsection{Real-World Shared Control for Tool Use}
\label{sec:real_shared_control}
We next evaluate whether the same controller assists human operators in real-world long-horizon tool use teleoperation.
Unlike the simulated corruptions, these tasks contain natural user errors, tool--environment contact, in-hand adjustment, and functional grasp transitions.
We evaluate six tasks: \textsc{Brush Sweep} (maintain a brush grasp while sweeping), \textsc{Screw Nut} (align and rotate a nut), \textsc{Hammer} (grasp and strike with impact), \textsc{Screwdriver} (grasp a screwdriver, insert it into a base, and rotate), \textsc{Spatula Serve} (scoop a hamburger with a spatula and place it on a plate), and \textsc{Flip Toast} (flip a toast slice with a spatula).

\begin{figure*}[t]
\centering
\includegraphics[width=\textwidth]{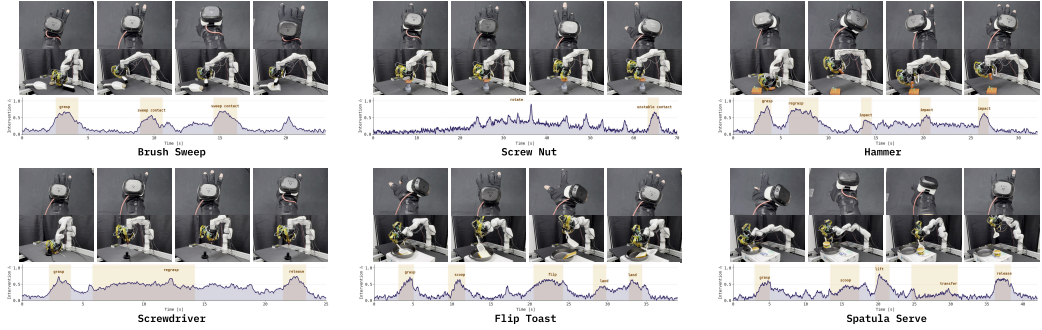}
\caption{\textbf{Selective intervention during real-world tool use.}
We visualize intervention magnitude $I_t=\|\hat u_t-u_t\|_2/(\|u_t\|_2+\epsilon)$ along representative rollouts. \ourmethod{} keeps intervention low during benign user motion and produces peaks near contact-critical moments such as release, impact, unstable tool contact, and regrasp.}
\label{fig:real_teleop_figure}
\vspace{-3mm}
\end{figure*}
\noindent\textbf{Reliability across tool-use tasks.}
Table~\ref{tab:real_teleop_results} shows that \ourmethod{} more than doubles raw teleoperation's average SR (29.2 to 78.3) and roughly doubles NHT (0.42 to 0.85), with a 25.8-point SR margin over DexGen.
\ourmethod{} \textsc{w/o Tactile} also improves over DexGen (52.5 to 59.2 SR), suggesting that consequence-aware selection helps beyond prior-only correction, while the full model gains another 19.1 points by using tactile state.
The gains are largest on contact-critical tasks such as \textsc{Spatula Serve} and \textsc{Flip Toast}, where premature release or unstable support frequently causes failure.
The improvement is not explained by uniformly slowing down execution: \ourmethod{}'s average CT (32.0\,s) lies within the inter-method range of 27.5 to 33.8\,s, while substantially improving success and holding stability.

\noindent\textbf{Selective intervention.}
Fig.~\ref{fig:real_teleop_figure} shows that \ourmethod{} does not continuously override the operator.
Intervention remains small during benign alignment and transport, but rises near contact-critical events such as release, impact, scoop, and regrasp.
Thus, \ourmethod{} improves reliability by selectively repairing low-level contact failures while leaving high-level task execution to the user.

\subsection{Low-Level Interface for Policy Learning}
\label{sec:policy_learning}

\begin{figure}[t]
\centering
\includegraphics[width=\textwidth]{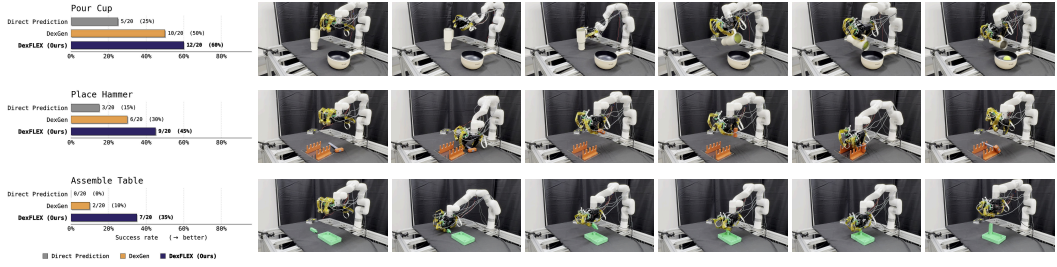}
\caption{\textbf{Policy learning with different action interfaces.}
All methods use the same image-based Flow-matching policy and 50 demos per task. Direct joint-action prediction is compared with fingertip-motion prediction executed by a prior-only DexGen controller and by \ourmethod{}.}
\label{fig:policy_learning}
\vspace{-3mm}
\end{figure}

Finally, we test whether \ourmethod{} can serve as a frozen low-level interface for visuomotor policy learning.
We evaluate three contact-rich tasks: \textsc{Pour Cup} (maintain a handle grasp while pouring), \textsc{Place Hammer} (grasp, reorient, and place into a holder), and \textsc{Assemble Table} (grasp, reorient, insert, and screw until lit).
For each task, we collect 50 demos and train the same Flow-matching policy~\citep{lipman2022flow}.
The methods differ only in the action interface: direct joint-command prediction, fingertip-motion prediction executed by a DexGen-style prior-only controller, or fingertip-motion prediction routed through the frozen \ourmethod{} module.
The DexGen-style baseline follows~\cite{yin2025dexteritygen}; reproduction details are provided in Appendix.
Each policy is evaluated for 20 rollouts per task.

Fig.~\ref{fig:policy_learning} shows that \ourmethod{} achieves the highest SR on all three tasks, reaching 46.7\% average SR compared with 13.3\% for direct joint actions and 30.0\% for DexGen.
With only 50 demonstrations per task, direct joint-action policies are brittle because the policy must learn both task intent and contact-sensitive execution in a high-dimensional joint space.
Predicting fingertip-motion drafts reduces this burden by letting the image policy express local task intent, while the low-level controller maps the draft back onto a contact-consistent dexterous motion distribution.
A prior-only controller provides this projection only through motion plausibility, whereas \ourmethod{} additionally uses predicted contact consequence to choose among plausible executions.
This improves policy success by 3.5$\times$ over direct joint actions and 1.56$\times$ over DexGen.
\section{Conclusions}
\label{sec:conclusions}
\vspace{-2mm}
We presented \ourmethod{}, a contact-aware foundation controller for command-guided dexterity.
\ourmethod{} refines upstream fingertip-motion drafts through a propose--predict--select procedure that combines state-conditioned motion proposals, latent contact-consequence prediction, and consequence-aware selection.
Across degraded-command simulation, real-world shared control, and visuomotor policy learning, the same trained controller improves contact-consistent execution without retraining, making it a reusable low-level interface beneath different command sources.

\textbf{Limitations and future work.}
\ourmethod{} currently assumes a fixed robot embodiment, tactile sensing, and a shared fingertip-motion command space. 
It can still fail under tactile sim-to-real gaps or when the upstream draft lies outside the support of the learned prior.
Future work will study cross-embodiment transfer, tactile adaptation, longer-horizon contact prediction, and tighter coupling between high-level policy learning and low-level contact-consistent execution.

\clearpage


\bibliography{reference}  




\end{document}